\documentclass[nohyperref]{article}

\usepackage{microtype}
\usepackage{graphicx}
\usepackage{subfigure}
\usepackage[normalem]{ulem}
\usepackage{booktabs} 

\usepackage{hyperref}

\usepackage[accepted]{icml2022}

\usepackage{amsmath}
\usepackage{amssymb}
\usepackage{mathtools}
\usepackage{amsthm}

\usepackage[capitalize,noabbrev]{cleveref}

\theoremstyle{plain}

\theoremstyle{definition}

\theoremstyle{remark}

\usepackage[textsize=tiny]{todonotes}

\newcommand{\transpose}     {\mathsf{T}}

\icmltitlerunning{Adaptively intrusive ROM closure}

\begin{document}

\onecolumn
\icmltitle{Continuous Methods : Adaptively intrusive reduced order model closure}



\icmlsetsymbol{equal}{*}

\begin{icmlauthorlist}
\icmlauthor{Emmanuel Menier}{1,2}
\icmlauthor{Michele Alessandro Bucci}{1}
\icmlauthor{Mouadh Yagoubi}{2}
\icmlauthor{Lionel Mathelin}{3}
\icmlauthor{Thibault Dairay}{4}
\icmlauthor{Raphael Meunier}{4}
\icmlauthor{Marc Schoenauer}{1}
\end{icmlauthorlist}

\icmlaffiliation{1}{TAU, Inria / LISN, Université Paris-Saclay / CNRS, Orsay, France}
\icmlaffiliation{2}{IRT - SystemX, Palaiseau, France}
\icmlaffiliation{3}{LISN, Université Paris-Saclay \& CNRS, Orsay, France}
\icmlaffiliation{4}{Michelin R\&D, Clermont Ferrand, France}
\icmlcorrespondingauthor{Emmanuel Menier}{emmanuel.menier@inria.fr}

\icmlkeywords{Neural ODE, Deep Learning, Reduced Order Models, Control}

\vskip 0.3in



\printAffiliationsAndNotice{} 

\begin{abstract}
Reduced order modeling methods are often used as a mean to reduce simulation costs in industrial applications. Despite their computational advantages, reduced order models (ROMs) often fail to accurately reproduce complex dynamics encountered in real life applications. To address this challenge, we leverage NeuralODEs to propose a novel ROM correction approach based on a time-continuous memory formulation. Finally, experimental results show that our proposed method provides a high level of accuracy while retaining the low computational costs inherent to reduced models. 
\end{abstract}

\section{Introduction}
\label{introduction}

The computational cost of classical simulation methods often constitutes a major impediment on quick and efficient design processes. Because of this, dimensionality reduction methods such as Proper Orthogonal Decomposition (also known as PCA) are often used to try to alleviate the computational costs of numerical simulations. A large body of literature is dedicated to this topic, as numerous methods have been proposed \cite{DMD,LoiseauSINDy,CROM,RowleyReview}. In this work, we focus on a specific application of the POD method, called POD-Galerkin, which is based on the identification of a low dimensional space on which both the solutions and governing partial differential equations (PDE) can be projected. 

The main advantage of the POD-Galerkin approach is that it retains physical information about the problem, as the equation is simply projected on a pre-defined truncated basis. This is in stark contrast with other approaches aiming to replace the physical equations  with physics-agnostic equation models \cite{POD-LSTM,SINDy}. Indeed, the conservation of dynamical information through the projection of the governing equations allows the ROM to predict dynamics in unseen conditions as it does not require any pre-emptive model fitting other than the computation of principal components. Despite this advantage, the POD-Galerkin approach has shown limitations when applied to non-linear PDE problems such as the Navier-Stokes equations \cite{NoackClosure}. In an effort to improve the performance of Galerkin reduced order models in such cases, the CD-ROM approach \cite{CD-ROM} has recently been developed and proposes to add a neural closure term to classical POD-Galerkin reduced order models to account for the loss of information inherent to linear dimensionality reduction.

The approach leverages Neural Ordinary Differential Equations (Neural ODEs, \citet{NeuralODE}) to learn a time-continuous, non-Markovian correction model for POD-Galerkin models. Using a continuous memory formulation inspired from theoretical results \cite{MoriZwanzig}, the model is able to retain past states of the system and retrieve information not readily available in the low dimensional modeling space (see Takens theorem, \citet{Takens}).

In this paper, we apply the CD-ROM approach to an industrial modeling problem. We show that the approach provides an adjustable degree of intrusivity as it can be used to learn a closure model as well as specific terms in partial differential equations. By augmenting an incomplete reduced order model of the problem at hand with the CD-ROM architecture, an efficient and flexible model is obtained, able to accurately simulate the problem in conditions unseen during training.

\section{Proposed modeling approach}\label{sec:method}
\subsection{Industrial problem : Rubber calendering process}

\begin{figure}[h]
    \centering
    \includegraphics[height=0.165\textheight]{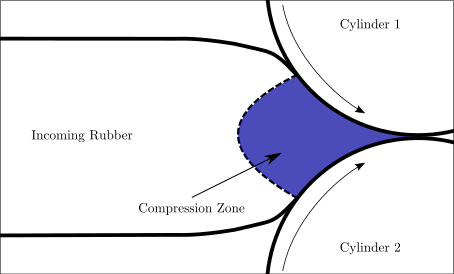}
    \includegraphics[height=0.165\textheight]{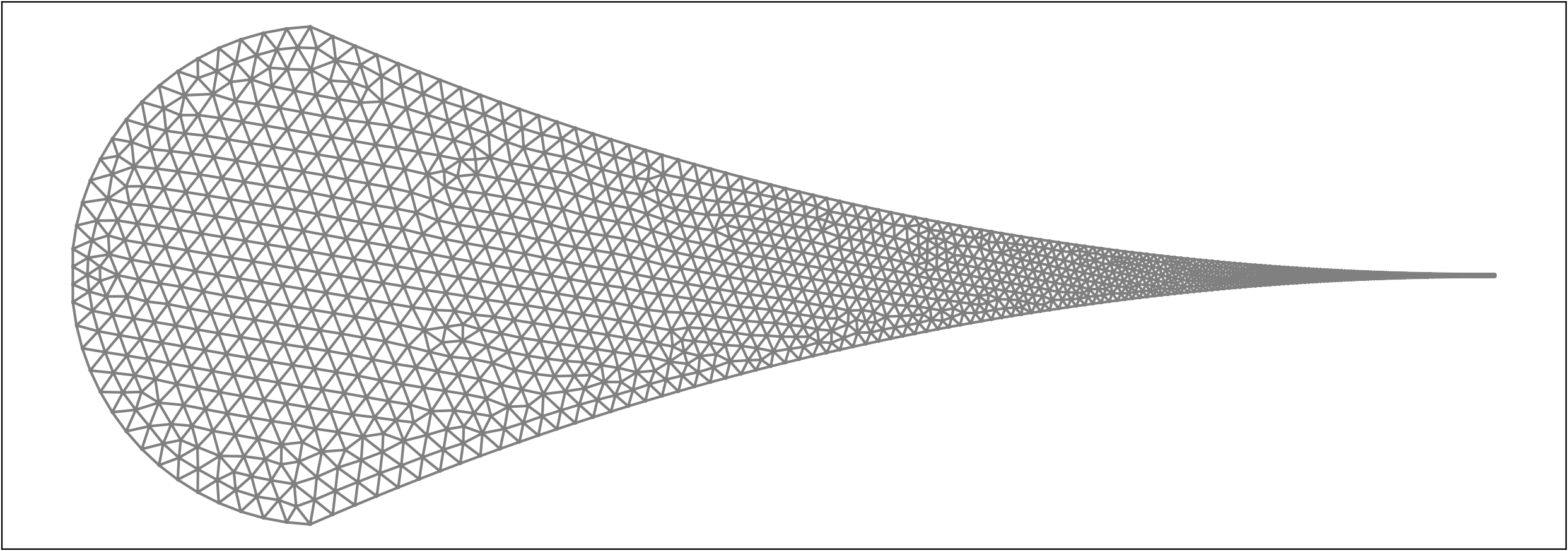}
    \caption{\textit{Left} : Schematic of the calendering process. \textit{Right} : Simulation mesh used in the finite element solver.}
    \label{fig:calendering_proces}
\end{figure}

This work focuses on the modeling of the rubber calendering process. Calendering is a manufacturing process which consists in passing a rubber sheet between two rollers to determine the thickness and mechanical properties of the material (see Figure \ref{fig:calendering_proces}). Because of the compression between the rotating cylinders, the rubber can heat up and deteriorate. To address this issue, one needs to estimate the heat generation inside the material under different cylinder rotation speeds in order to determine acceptable process conditions. The issue is that simulating the problem in a classical finite elements solver can take too much time, limiting the applicability of full order simulation approaches to the control of the process. We hence propose to use model order reduction to lower the cost of simulating the problem.

\subsection{Governing equations}

In this work, we focus on the modeling of the dynamics of the temperature field, as we have observed that the velocity field $\mathbf{u}$ can be inferred with sufficient accuracy from the cylinder rotational speed ($\mathcal{S}$) and the temperature field ($T$) such that $\mathbf{u} = \mathbf{u}(T,\mathcal{S})$, since it is in a quasi-steady regime due to the high viscosity of the flow. Despite this simplification, the reduction of the temperature dynamics is challenging for classical methods such as the POD-Galerkin approach. To understand this, one needs to look at the PDE governing the temperature dynamics:

\begin{equation}
    \frac{\partial T}{\partial t} = \frac{\lambda}{\rho C_p} \nabla^2 T - \mathbf{u}\cdot\nabla T + \frac{\eta(\mathbf{u},T)\gamma(\mathbf{u})^2}{\rho C_p}
\label{eq:TrueDynamics}
\end{equation}

where $T$ represents the value of the temperature at any mesh point, $\mathbf{u}$ the velocity, $\eta$ the dynamic viscosity of the material and $\gamma$ the deformation rate.  Because of the linear nature of the Laplacian and gradient operators, the first two terms of the dynamics can easily be projected on a linear low dimensional basis of spatial modes, reducing their computation to simple tensorial operations as explained in \citet{Noack,CD-ROM}. The last term however, is a source term accounting for the heat generated by the deformation of the rubber. This phenomenon is strongly nonlinear and cannot be reduced linearly. Computing its reduced form would require back and forth exchanges between the full order solver and the reduced model, which would directly impact the computational performances of the ROM. To avoid these costly steps, we extend the CD-ROM approach to model the last term of Eq.~\eqref{eq:TrueDynamics} in addition to the required correction term.

\subsection{ROM formulation}

\begin{figure}[ht]
    \centering
    \includegraphics[width = 0.9\textwidth]{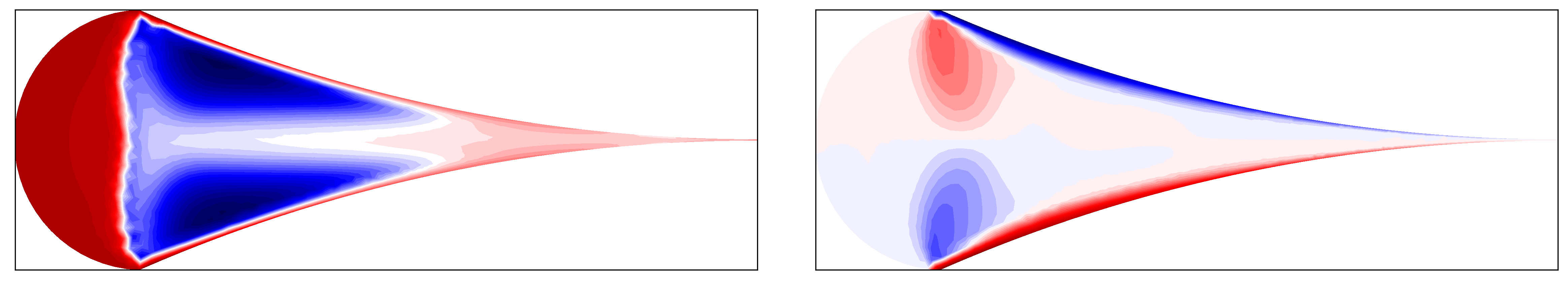}
    \caption{\textit{Left}: Leading temperature POD mode. \textit{Right}: Vertical component of the leading velocity POD mode. Note : The legend is not reported as the point values of the field are not relevant, this is because the POD bases are normalised so that $V_x^\transpose V_x = I$. }
    \label{fig:ExampleModes}
\end{figure}

To construct the ROM, we first assemble a collection of solutions of the system at different time steps and using different cylinder rotational speeds by simulating Eq.~\eqref{eq:TrueDynamics} with the finite element solver MEF++\footnote{MEF++ --- Wikipédia,  \url{http://fr.wikipedia.org/w/index.php?title=MEF\%2B\%2B\&oldid=192108614}} \cite{MefPaper}. By computing the POD of the obtained dataset, we can extract a reduced number of principal components (or modes) optimally approximating the data. Following this strategy, we compute two orthonormal bases of modes, $V_T \in \mathbb{R}^{n_c \times n_T}$ for the temperature and $V_\mathbf{u} \in \mathbb{R}^{2n_c \times n_\mathbf{u}}$ for the velocity. Here $n_c$ denotes the number of grid cells in the mesh, $n_T$ the number of selected temperatures modes, and $n_\mathbf{u}$ the number of selected velocity modes, so that each column of the reduced bases matrix $V_T$ and $V_\mathbf{u}$ represents a complete field. To better illustrate the idea, Figure~\ref{fig:ExampleModes} displays the leading mode of each basis. 

After computing these two bases, one can approximate the solution as linear combinations of the principal components: $T = V_T \alpha_T$ and $\mathbf{u} = V_\mathbf{u} \alpha_\mathbf{u}$. With this formulation, solving the problem reduces to computing the low dimensional vectors of POD coordinates $\alpha_T \in \mathbb{R}^{n_T}$ and $\alpha_\mathbf{u} \in \mathbb{R}^{n_\mathbf{u}}$. As explained above, the velocity field is directly evaluated from the cylinder rotational speed and the temperature field, leaving only the dynamics of the temperature reduced coordinates $\alpha_T$ to be modeled. Following the POD-Galerkin method, the reduced forms of the temperature and velocity fields are injected in the temperature dynamics (Eq.~\eqref{eq:TrueDynamics}), which are then projected on the temperature POD basis $V_T$, yielding a system of $n_T$ ordinary differential equations:

\begin{equation}
    \frac{d \alpha_T}{dt} = \underbrace{
    \frac{\lambda}{\rho C_p} V_T^\transpose\nabla^2 V_T \alpha_T - \alpha_\mathbf{u} V_T^\transpose V_\mathbf{u} \cdot \nabla V_T \alpha_T}_{\mathcal{R}(\alpha_T,\alpha_\mathbf{u},\mathcal{S})} + \underbrace{
    V_T^\transpose \frac{\eta(V_\mathbf{u}\alpha_\mathbf{u},V_T\alpha_T)\gamma(V_\mathbf{u}\alpha_\mathbf{u})^2}{\rho C_p}}_{\mathcal{I}(\alpha_T,\alpha_\mathbf{u},\mathcal{S})}
    \label{eq:reduced_system}
\end{equation}

The number of POD modes needed to accurately reconstruct the temperature field is very low ($\sim 5$), which means the reduced system in Eq.~\eqref{eq:reduced_system} can be simulated very quickly, achieving significant computational cost reduction \textit{w.r.t.} full order solvers. However, as introduced above, the equation can be separated in two parts: a reducible part $\mathcal{R}$ which easily expresses in terms of the reduced coordinates $\alpha$, and an irreducible part $\mathcal{I}$ which cannot be directly evaluated in the reduced space. This would normally be a major impediment on the use of model order reduction methods to solve this problem. However, using the CD-ROM approach, we can learn the effect of $\mathcal{I}$ on the reduced dynamics using a neural network. Allowing for the extension of reduced order modeling approaches to previously irreducible problems, while retaining as much as possible from the original dynamical equations:

\begin{align}
    \frac{d \alpha_T}{dt}(t) &= \mathcal{R}(\alpha_T,\alpha_\mathbf{u},\mathcal{S}) + \mathcal{NN}(\alpha_T,\alpha_\mathbf{u},\mathcal{S},y) \label{eq:cdrom_form}\\
    y(t) & = \int_{-\infty}^t e^{(s-t)\Lambda} x(s) \mathrm{d}s, \qquad x(t) = [\alpha_T(t), \mathcal{S}(t)] 
\end{align}

where $\Lambda$ is a positive diagonal matrix corresponding to the time horizon matrix defined in \citet{CD-ROM} and $y(t)$ is a memory term specifically designed to be continuously integrable in parallel with the reduced dynamics as a simple linear system. In \citet{CD-ROM} the critical role of the memory term $y(t)$ in retrieving information necessary to the correction of reduced order models is underlined. Using this architecture in combination with Neural ODEs, we show below that the neural network in Eq.~\eqref{eq:cdrom_form} can be trained in an \textit{a posteriori} fashion to appropriately correct the incomplete reduced model $\mathcal{R}$.

\section{Results}

\subsection{Experiment setup}

To train the parameters of the closure model, we simulate the system with different cylinder rotational speed trajectories defined as a sum of sines with random coefficients:

\begin{equation}
    \mathcal{S}(t) = c_0 + \sum_{i=4}^{7} c_i \sin\left(\frac{2\pi t}{2^i}\right), \qquad c_0 \sim \mathcal{N}(1,0.25), \quad c_i \sim \mathcal{N}(0,0.25)
\end{equation}

By sampling 20 trajectories $\mathcal{S}(t)$ from this distribution, we can simulate the system under varying conditions and apply the closure learning strategy described in the previous section. After computing the two POD bases $V_\mathbf{u}$ and $V_T$, we select the leading $n_\mathbf{u} = 4$ and $n_T = 6$ velocity and temperature POD modes, which represent more than $95 \%$ of the variance in the dataset. Finally, the relation $\alpha_\mathbf{u} = f(\alpha_T,\mathcal{S})$ is approximated using ridge regression. This choice is made to reduce as much as possible the variance in $\mathcal{R}$, while the approximation error introduced by this simple model is corrected by the neural closure model $\mathcal{I} = \mathcal{NN}(\alpha_T,\alpha_\mathbf{u},\mathcal{S},y)$.

\subsection{Test performance}

The model is trained using $80\%$ of the dataset using NeuralODEs in combination with the Adaptive Checkpointing Adjoint method \cite{ACA}. The objective $\mathcal{L}$ is defined as the mean squared Euclidean distance between the simulated reduced coordinates and their true value $\alpha_T^\star$:

\begin{equation}
    \mathcal{L} = \frac{1}{n_t+1} \sum_{i=0}^{n_t} \left\| \alpha_T^\star(t_i) - \alpha_T(t_i) \right\|_2^2
\end{equation}

\begin{figure}[h]
    \centering
    \includegraphics[width=\textwidth]{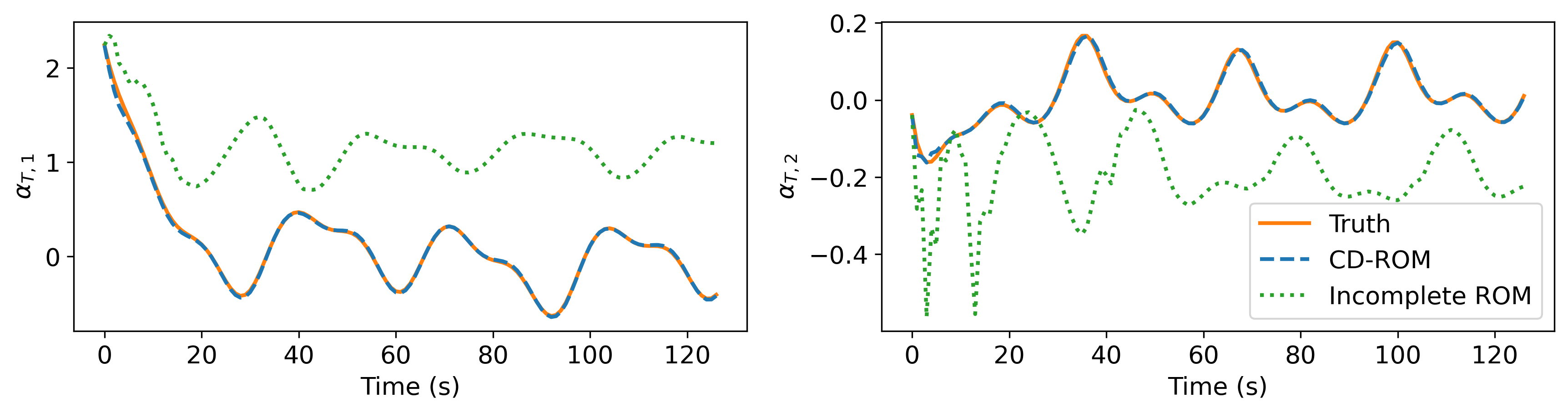}
    \caption{Performance of the corrected ROM (Eq.~\eqref{eq:cdrom_form}) on a trajectory unseen during training. \textit{Left}: Trajectory of the first mode. \textit{Right}: Trajectory of the second mode. The incomplete ROM ($\mathcal{R}$ in Eq.~\eqref{eq:reduced_system}) is also shown for comparison.}
    \label{fig:TestResults}
\end{figure}

The remaining $20 \%$ of the dataset is used for testing. Figure \ref{fig:TestResults} presents the performance of the model and its uncorrected counterpart on a test trajectory. We also compute the RMSE normalized by the standard deviation of the data to provide a quantitative indication of the performance of the model on the test trajectories. Obtained results show that the CD-ROM trajectory fits the true trajectory almost perfectly, compared with the incomplete ROM (Figure \ref{fig:TestResults}). The final NRMSE computed over the complete test set is of $2.5 \%$ . Moreover, the simulation of the corrected reduced model is much more computationally efficient than finite element solvers as the parallel simulation of 128 trajectories only takes a few seconds on a RTX 2080 GPU, while the simulation of a single trajectory in our finite elements solver took about 5 minutes. Note that these simulation times simply provide a rough estimation of the performance gap, as they heavily depend on the hardware, implementation and simulation parameters of both the ROM and the FE model.

\section{Conclusion}

In this work, we demonstrate the interest of using neural networks for the modeling of dynamical systems. By creating a hybrid model combining physical knowledge and a neural closure term, we obtain a continuous model able to simulate a complex physical problem efficiently.

We demonstrate that this model can be used outside of its training conditions while retaining a high level of accuracy. Our proposal underlines the adaptability of the CD-ROM approach to ill-posed reduction problems such as those involving highly nonlinear terms, while retaining a high degree of interpretability compared to physics-agnostic NeuralODEs. In the future, the ability of the model to generalise to new materials in addition to new conditions should be investigated. The proposed approach could then be used in model predictive control strategies to optimally tune the parameters of the manufacturing process.


\bibliography{example_paper}
\bibliographystyle{icml2022}


\end{document}